\documentclass[runningheads,a4paper]{llncs}

\usepackage{amssymb}
\usepackage{graphicx}
\usepackage{url}
\urldef{\mailsa}\path|wangy@cse.unsw.edu.au|
\urldef{\mm}\path|lin.wu@uq.edu.au|

\usepackage{latexsym}
\usepackage{epsfig}
\usepackage{color}
\usepackage{footmisc}
\usepackage[lined,ruled,vlined]{algorithm2e}
\usepackage{enumerate}
\usepackage{graphicx}
\usepackage{amsmath}
\usepackage{wrapfig}
\usepackage{multirow}

\newcommand{\etal}{\emph{et al. }}

\newcommand{\keywords}[1]{\par\addvspace\baselineskip
\noindent\keywordname\enspace\ignorespaces#1}

\begin{document}

\mainmatter  

\title{Finding Modes by Probabilistic Hypergraphs Shifting}

\titlerunning{ }

\author{%
    Yang Wang$^{\dag}$ \and
    Lin Wu$^{\ddag}$
    }
    \authorrunning{Yang Wang, Lin Wu}
\institute{$^{\dag}$The University of New South Wales, Sydney, Australia\\
\mailsa \\
\vspace{2mm}
${^\ddag}$The University of Queensland, Brisbane, Australia\\
\mm\\
}

%
%

\maketitle

\begin{abstract}
In this paper, we develop a novel paradigm, namely hypergraph shift, to find robust graph modes by probabilistic voting strategy, which are semantically sound besides the self-cohesiveness requirement in forming graph modes.
Unlike the existing techniques to seek graph modes by shifting vertices based on pair-wise edges (i.e, an edge with $2$ ends), our paradigm is based on shifting high-order edges (hyperedges) to deliver graph modes. Specifically, we convert the problem of seeking graph modes as the problem of seeking maximizers of a novel objective function with the aim to generate good graph modes based on sifting edges in hypergraphs. As a result, the generated graph modes based on dense subhypergraphs may more accurately capture the object semantics besides the self-cohesiveness requirement.
We also formally prove that our technique is always convergent. Extensive empirical studies on synthetic and real world data sets are conducted on clustering and graph matching.
They demonstrate that our techniques significantly outperform the existing techniques.
\end{abstract}

\section{Introduction}

Seeking graph based modes is of great importance to many applications in machine learning literature, e.g., image segmentation \cite{Kim2011NIPS}, feature matching \cite{Cho12CVPR}. In order to find the good modes of graphs,
Pavan \etal \cite{Dominant} converted the problem of mode seeking into the problem of discovering dense subgraphs, and proposed a constrained optimization function for this purpose.
Liu \etal \cite{GraphShift} proposed another method, namely graph shift. It generalized the idea of non-parametric data points shift paradigms (i.e., Mean Shift \cite{MeanShift} and Medoid Shift \cite{MedoidShift,Geometricshift,Medianshift} to graph shift for graph mode seeking). An iterative method is developed to get the local maximizers, of a constrained objective function, as the good modes of graphs.
\begin{figure}[hbt]
\begin{center}
\begin{tabular}{c}
\epsfxsize=80mm\epsfbox{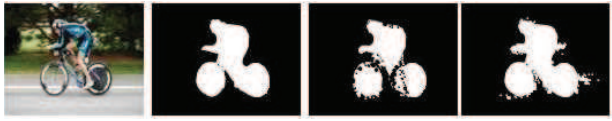}\\
\end{tabular}
\caption{Comparison between graph shift and hypergraph shift on saliency detection, from left to right: input image, ground truth, graph shift, hypergraph shift.}
\label{fig:intro-graph-hyper}
\end{center}
\end{figure}
While the graph (vertices) shift paradigm may deliver good results in many cases for graph mode seeking, we observe the following limits.
Firstly, the graph modes generated based on shifting vertices only involve the information of pair-wise edges between vertices. As a result, the generated graphs modes
may not always be able to precisely capture the overall semantics of objects.
Secondly, the graph shift algorithm is still not strongly robust to the existence of a large number of outliers.
Besides, no theoretical studies are conducted to show the convergence of iteration of shifting.

\textbf{Our Approach}: Observing the above limits, we propose a novel paradigm, namely hypergraph shift, aimed at generating graph modes with high order information. Different from graph shift paradigms that only shift vertices of graphs based on pair-wise edges, our technique shifts high order edges (hyperedges in hypergraphs).
Our technique consists of three key phases,  1) mode seeking (section ~\ref{sec:modeseek}) on subhypergraphs, 2) probabilistic voting (section ~\ref{sec:voting}) to determine a set of hyperedges to be expanded in mode seeking, and 3) iteratively perform the above two stages until convergence.

By these three phases, our approach may accurately capture the overall semantics of objects. Fig.~\ref{fig:intro-graph-hyper} illustrates an example where the result of our approach for hypergraph shift can precisely capture the the scene of a person riding on a bicycle. Nevertheless, the result performed by graph shift method in \cite{GraphShift} fails to capture the whole scene; instead, by only focusing on the requirement of self-cohesiveness, three graph modes are generated.

\textbf{Contributions}: To the best of our knowledge, this is the first work based on shifting hyperedges to conduct graph mode seeking. Our contributions may be summarized as follows. (1) We specify the similarities on hyperedges, followed by an objective function for mode seeking on hypergraphs. (2) An effective hypergraph shift paradigm is proposed. Theoretical analysis for hypergraph shift is also provided to guarantee its convergence. The proposed algorithm is naturally robust to outliers by expanding modes via the probabilistic voting strategy.
(3) Extensive experiments are conducted to verify the effectiveness of our techniques over both synthetic and real-world datasets.

\textbf{Roadmap}: We structure our paper as follows: The preliminaries regarding hypergraph are introduced in section \ref{sec:notation}, followed by our technique for hypergraph shift
in sections \ref{sec:mode} and \ref{sec:shift}.
Experimental studies are performed in section \ref{sec:exp}, and we conclude this paper in section \ref{sec:con}.

\section{Probabilistic Hypergraph Notations}\label{sec:notation}
Different from simple graph, each edge of hypergraph (known as hyperedge) can connect more than two vertices. Formally, we denote a weighted hypergraph as $\mathbf{G} = (\mathcal{V}, \mathcal{E}, \mathcal{W})$, with vertex set as $\mathcal{V}=\{v_1, v_2,\ldots,v_{|\mathcal{V}|}\}$,
hyperedge set as $\mathcal{E}=\{e_1,e_2,\ldots,e_{|\mathcal{E}|}\}$, and
$\mathcal{W} = \{w(e_1),w(e_2),\cdot\cdot\cdot,w(e_{|\mathcal{E}|})\} $, where $w(e_i)$ is the weight of $e_i$.
The relationship between the hyperedges and vertices is defined by incidence matrix $\mathbf{H} \in \mathbb{R}^{|\mathcal{V}|\times |\mathcal{E}|}$.
Instead of assigning a vertex $v_i$ to a hyperedge $e_j$ with a binary decision,
we establish the values probabilistically \cite{DingSegment,HuangCVPR2010}. Specifically, we define the entry $h_{v_i,e_j}$ of $\mathbf{H}$ as Eq.~\eqref{eq:incidencedef}.
\begin{equation}\label{eq:incidencedef}
h_{v_i,e_j}=\left\{
  \begin{array}{ll}
    p(v_i|e_j), & \hbox{if $v_i\in e_j$;} \\
    0, & \hbox{otherwise.}
  \end{array}
\right.
\end{equation}
where $p(v_i|e_j)$ describes the likelihood that a vertex $v_i$ is connected to hyperedge $e_j$.
Then we define a diagonal matrix $\mathbf{D}_e$ regarding the degree of all hyperedges,
with $\mathbf{D}_{e}(i,i)=\delta(e_i)=\sum_{v\in \mathcal{V}}h_{v,e_i}$,
and a diagonal matrix $\mathbf{D}_v$ regarding the degree of all vertices, with  $\mathbf{D}_{v}(i,i)=\sum_{e \in \mathcal{E}}h_{v_i,e}w(e)$. Based on that,
to describe the similarity between hyperedges, we define a novel \textbf{\textsl{hyperedge-adjacency matrix}} $\mathbf{M}\in \mathbb{R}^{|\mathcal{E}|\times |\mathcal{E}|}$ in the context of hypergraph. Specifically, we have
\begin{equation}\label{eq:hyper-adjacency}
\textbf{M}(i,j)= \left\{
\begin{array}{ll}
w(e_i)\frac{|e_i \cap e_j|}{\delta(e_i)} + w(e_j)\frac{|e_i \cap e_j|}{\delta(e_j)} & \hbox{$i \neq j$}\\
0, & \hbox{otherwise}
\end{array}
\right.
\end{equation}

\begin{figure}[hbt]
\begin{center}
\begin{tabular}{c}
\epsfxsize=100mm\epsfbox{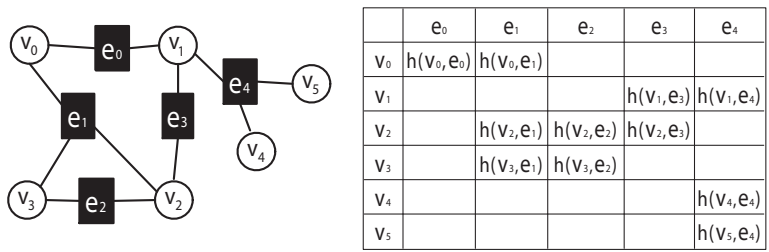}
\end{tabular}
\caption{A toy example on hypergraph. Left: a hypergraph. Right: The incidence matrix of hypergraph.}\label{fig:DFS-tree}
\end{center}
\end{figure}

\begin{example}
Consider the case in Fig.\ref{fig:DFS-tree}, for $e_2$ and $e_3$, the only common vertex is $v_2$, then, we have $|e_2 \cap e_3|$=1, and the affinity value between $e_2$ and $e_3$ is $\textbf{M}(2,3)=w(e_2)\cdot\frac{1}{2}+w(e_3)\cdot\frac{1}{2}=\frac{w(e_2)+w(e_3)}{2}$.
\end{example}
Now, we describe the modes of hypergraph.

\section{Modes of Hypergraph}\label{sec:mode}

We consider the mode of a hypergraph as a dense subhypergraph consisting of hyperedges with high self-compactness. We first define the \textbf{hypergraph density}, then formulate the modes of a hypergraph, which leads to our hypergraph shift algorithm in section \ref{sec:shift}.

{\bf Hypergraph Density.} We describe hypergraph $\mathbf{G}$ with $n$ hyperedges by probabilistic coordinates fashion as $\mathbf{p} \in \Delta^{n}$, where $\Delta^{n} = \{\mathbf{p}|\mathbf{p} \geq 0, |\mathbf{p}|_1 = 1 \}$, $|\mathbf{p}|_1$ is the $L_1$ norm of vector $\mathbf{p}$, and $\mathbf{p}=\{\mathbf{p}_1,\mathbf{p}_2,\ldots,\mathbf{p}_n\}$. Specifically, $\mathbf{p}_i$ indicates the probability of $e_i$ contained by the probabilistic cluster of $\mathbf{G}$.
Then the affinity value between any pair-wise $\mathbf{x} \in \Delta^{n}$ and $\mathbf{y} \in \Delta^{n}$ is defined as $\mathbf{m}(\mathbf{x},\mathbf{y})=\sum_{i,j} \mathbf{M}(i,j) \mathbf{x}_i \mathbf{y}_j = \mathbf{x}^T \mathbf{M} \mathbf{y}$.
The \textbf{hypergraph density} or self-cohesiveness of $\mathbf{G}$, is defined as Eq.~\eqref{eq:quadratic}.
\begin{equation}\label{eq:quadratic}
\mathcal{F}(\mathbf{p}) = {\mathbf{p}}^T \mathbf{M} \mathbf{p}.
\end{equation}
Intuitively, \textbf{hypergraph density} can be interpreted by the following principle. Suppose hyperedge set $\mathcal{E}$ is mapped to $\mathbf{I}=\{i_m | m=1,\ldots,|\mathcal{E}|\}$, which is the representation in a specific feature space regarding all hyperedges in $\mathcal{E}$, where we define a kernel function $\mathcal{K}: \mathbf{I} \times \mathbf{I} \rightarrow \mathbb{R}$. Specifically, $\mathcal{K}(i_m,i_n) = \mathbf{M}(m,n)$. Thus, the probabilistic coordinate $\mathbf{p}$ can be interpreted to be a probability distribution, that is, the probability of $i_m$ occurring in a specific subhypergraph is $\mathbf{p}_m$. Assume that the distribution is sampled $\mathcal{N}$ times, then the number of data $i_m$ is $\mathcal{N} \mathbf{p}_m$. For $i_m$, the density is $d(i_m) = \frac{\sum_n \mathcal{N} \mathbf{p}_m \mathcal{K}(m,n)}{\mathcal{N}}$, then we have the average density of the data set:
\begin{equation}\label{eq:density-ave}
\small
\overline{d}=\frac{\sum_m \mathcal{N} \mathbf{p}_m d(i_m)}{\mathcal{N}}=\sum_{m \neq n} \mathbf{p}_m \mathcal{K}(i_m,i_n) \mathbf{p}_n=\mathbf{p}^T \mathbf{M}\mathbf{p}
\end{equation}

\begin{definition}\label{def:mode}\textbf{(Hypergraph Mode)}
The mode of a hypergraph $\mathbf{G}$ is represented as a dense subhypergraph that locally maximizes the Eq.~\eqref{eq:quadratic}.
\end{definition}

Given a vector $\mathbf{p} \in \Delta^n$, the \textbf{\textsl{support}} of $\mathbf{p}$ is defined as the set of indices corresponding to its nonzero components:
$\theta(\mathbf{p})=\{i\in |\mathcal{E}|: \mathbf{p}_i\neq 0\}$.
Thus, its corresponding subhypergraph is $\mathbf{G}_{\theta(\mathbf{p})}$, composed of all vertices whose indices are in $\theta(\mathbf{p})$.
If $\mathbf{p}^*$ is a local maximizer i.e., the mode of $\mathcal{F}(\mathbf{p})$, then $\mathbf{G}_{\theta(\mathbf{p}^*)}$ is a dense subhypergraph.
Hence, the problem of mode seeking on a hypergraph is equivalent to maximizing the density measure function $\mathcal{F}(\mathbf{p})$,
which is taken as the criterion to evaluate the goodness of any subhypergraph.

To find the modes, i.e., the local maximizers of Eq.~\eqref{eq:quadratic}, we classify it into the standard quadratic program (StQP) \cite{Dominant,StQP}:
\begin{equation}\label{eq:mode}
\max \mathcal{F}(\mathbf{p}), s.t. \mathbf{p} \in \Delta^n,
\end{equation}
According to \cite{Dominant,StQP}, a local maximizer $\mathbf{p}^*$ meets the Karush-Kuhn-Tucker(KKT) condition. In particular,
there exist $n+1$ real Lagrange multipliers $\mu_i \geqslant 0 (1 \leq i \leq n)$ and $\lambda$, such that:
\begin{equation}\label{eq:KKT}
(\mathbf{M}\mathbf{p})_i -\lambda +\mu_i=0
\end{equation}
for all $i=1,\ldots,n$, and $\sum_{i=1}^n \mathbf{p}^*_i\mu_i=0$.
Since $\mathbf{p}^*$ and $\mu_i$ are nonnegative, it indicates that $i\in \theta(\mathbf{p}^*)$ implies $\mu_i=0$.
Thus, the KKT condition can be rewritten as:
\begin{equation}
(\mathbf{M}\mathbf{p}^*)_i \left\{
  \begin{array}{ll}
    =\lambda, & \hbox{$i\in \theta(\mathbf{p}^*)$;} \\
    \leqslant \lambda, & \hbox{otherwise.}
  \end{array}
\right.
\end{equation}
where $(\mathbf{M}\mathbf{p}^*)_i$ is the affinity value between $\mathbf{p}^*$ and $e_i$.

\section{Hypergraph Shift Algorithm}\label{sec:shift}
Commonly the hypergraph can be very large, a natural question is how to perform modes seeking on a large hypergraph? To answer this question, we perform mode seeking on subhypergraph, and determine whether it is the mode of the hypergraph. If not, we shift to a new subhypergraph by expanding the neighbor hyperedges of the current mode to perform mode seeking. Prior to that, we study the circumstances that determine whether the mode of a subhypergraph is the mode of that hypergraph.

Assume $\mathbf{p}_\mathcal{S}^*$ is the mode of subhypergraph $\mathcal{S}$ containing $m=|\theta(\mathbf{p}_\mathcal{S}^*)|$ hyperedges,
then we expand the $m$ dimensional $\mathbf{p}_\mathcal{S}^*$ to $|\mathcal{E}|$ dimensional $\mathbf{p}^*$ by filling zeros into the components, whose indices are in the set of $\mathbf{G}-\mathcal{S}$. Based on that, Theorem \ref{thm:mode-judge} is presented to determine whether $\mathbf{p}_\mathcal{S}^*$ is the mode of hypergraph $\mathbf{G}$.

\begin{theorem}\label{thm:mode-judge}
A mode $\mathbf{p}_\mathcal{S}^*$ of the subgraph $\mathcal{S}$ is also the mode of hypergraph $\mathbf{G}$ if and only if for all hyperedge $e_j$, $\mathbf{m}(\mathbf{p}^*, \mathbf{I}_j) \leqslant \mathcal{F}(\mathbf{p}^*)=\mathcal{F}_\mathcal{S}(\mathbf{p}_\mathcal{S}^*)$, $j\in \mathbf{G}-\mathcal{S}$, where $\mathbf{p}^*$ is computed from $\mathbf{p}_\mathcal{S}^*$ by filling zeros to the elements whose indices are in $\mathbf{G}-\mathcal{S}$ and $\mathbf{I}_j$ is the vector containing only hyperedge $e_i$ where its $i$-th element is 1 with others 0.

\rm
\noindent\textbf{Proof.} Straightforwardly, $\theta(\mathbf{p}^*) = \theta(\mathbf{p}_\mathcal{S}^*)$, $\mathcal{F}(\mathbf{p}^*) = \mathcal{F}_\mathcal{S}(\mathbf{p}_\mathcal{S}^*)$. Due to $\mathbf{m}(\mathbf{p}^*, \mathbf{I}_j) \leqslant
\mathcal{F}(\mathbf{p}^*) = \mathcal{F}_\mathcal{S}(\mathbf{p}_\mathcal{S}^*) = \lambda$, $\forall j \in \mathbf{G}-\mathcal{S}$, $\mathbf{p}^*$ is the mode of hypergraph $\mathbf{G}$. Otherwise if $\mathbf{m}(\mathbf{p}^*,\mathbf{I}_j) > \mathcal{F}(\mathbf{p}^*) = \lambda$, which indicates that $\mathbf{p}^*$ violates the KKT condition, thus it is not the mode of $\mathbf{G}$. $\hfill\blacksquare$
\end{theorem}

Next, we introduce our hypergraph shift algorithm, which consists of two steps: The first step performs mode seeking on an initial subhypergraph. If the mode obtained in the first step is not the mode of that hypergraph, it shifts to a larger subhypergraph by expanding the support of the current mode to its neighbor hyperedges using the technique, namely probabilistic voting. The above steps alternatively proceed until the mode of hypergraph is obtained.

\subsection{Higher-order Mode Seeking}\label{sec:modeseek}
Given an initialization of $\mathbf{p}(0)$, we find solutions of Eq.~\eqref{eq:mode} by using the replicator dynamics, which is a class of continuous and discrete-time dynamical systems arising in evolutionary game theory \cite{GameTheory95}. In our setting, we use the following form:
\begin{equation}\label{eq:replicator}
\mathbf{p}_i(t+1) = \mathbf{p}_i(t) \frac{(\mathbf{M} \cdot \mathbf{p}(t))_i }{\mathbf{p}(t)^T \mathbf{M} \mathbf{p}(t)}, i=1,\ldots,|\mathcal{E}|
\end{equation}
It can be seen that the simplex $\Delta^n$ is invariant under these dynamics, which means that every trajectory starting in $\Delta^n$ will remain in $\Delta^n$ for all future times. Furthermore, according to \cite{GameTheory95}, the objective function of Eq.~\eqref{eq:quadratic} strictly increases along any nonconstant trajectory of Eq.~\eqref{eq:replicator}, and its asymptotically stable points are in one-to-one with local solutions of Eq.~\eqref{eq:mode}.

\subsection{Probabilistic Voting}\label{sec:voting}

\begin{figure}[hbt]
\begin{center}
\begin{tabular}{c}
\epsfxsize=120mm\epsfbox{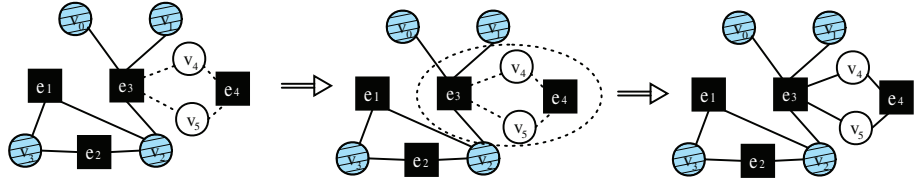}
\end{tabular}
\caption{Probabilistic voting strategy. The hyperedge $e_3$ is selected to be the dominant seed because of its higher closeness, as measured by Eq.~\eqref{eq:seed}.
We start the expansion from $e_3$ and then include its nearest neighbor $e_4$ into the mode.}
\label{fig:voting}
\end{center}
\end{figure}

\vspace{-2mm}
We propose to find the \textbf{\textsl{dominant seeds}} of the subhypergraph, from which we perform hypergraph shift algorithm.
Before presenting the formal definition of dominant seeds, we start with the intuitive idea that the assignment of hyperedge-weights induces an assignment of weights on the hyperedges.
Therefore, the average weighted degree of a hyperedge $e_k$ from subhypergraph $\mathcal{S}$ is defined as:
\begin{equation}\label{eq:awdeg}
g_{\mathcal{S}}(e_k)=\frac{1}{|\mathcal{S}|}\sum_{e_j\in \mathcal{S}} \mathbf{M}(k,j)
\end{equation}
Note that $g_{e_k}(e_k)$ = 0 for any $e_k \in \mathcal{S}$. Moreover, if $e_j \nsubseteq \mathcal{S}$, we have:
\begin{equation}
\psi_{\mathcal{S}}(e_i,e_j)=\mathbf{M}(i,j) - g_{\mathcal{S}}(e_i)
\end{equation}
Intuitively, $\psi_{\mathcal{S}}(e_i,e_j)$ measures the relative closeness between $e_j$ and $e_i$ with respect to the average closeness between $e_i$ and its neighbors in $\mathcal{S}$.

Let $\mathcal{S} \subseteq \mathcal{E}$ be a nonempty subset of hyperedges, and $e_i \in \mathcal{S}$. The weight of $e_i$ is given as
\begin{equation}\label{eq:close-measure}
w_{\mathcal{S}}(e_i)=
\left\{
  \begin{array}{cc}
    1, & \hbox{ if $|\mathcal{S}|$ =1;} \\
    \sum\limits_{e_j\in \mathcal{S}- \{e_i\}} \psi_{\mathcal{S}- \{e_i\}} (e_j,e_i) w_{\mathcal{S}- \{e_i\}}(e_j), & \hbox{otherwise.}
  \end{array}
\right.
\end{equation}
$w_{\mathcal{S}}(e_i)$ measures the overall closeness between hyperedge $e_i$ and other hyperedges of $\mathcal{S} - \{e_i\}$. Moreover, the total weight of $\mathcal{S}$ is defined as $W(\mathcal{S}) = \sum_{e_i \in \mathcal{S}}w_\mathcal{S}(e_i)$. Finally, we formally define the dominant seed of subhypergraph $\mathcal{S}$ as follows.
\begin{definition}\label{def:seed}\textbf{(Dominant Seed)}
The dominant seed of a subhypergraph $\mathcal{S}$ is the subset of hyperedges with higher closeness than others.
\end{definition}
Besides, the closeness of the dominant seed is evaluated as follows:
\begin{equation}\label{eq:seed}
p(e_i|\mathcal{S})=
\left\{
 \begin{array}{cc}
 \frac{w_{\mathcal{S}}(e_i)}{W(\mathcal{S})}, & \hbox{ if $e_i \in \mathcal{S}$}\\
 0, & \hbox{otherwise.}
 \end{array}
\right.
\end{equation}
We utilize dominat seeds to expand the current subhypergraph, which is named \textbf{\textsl{probabilistic voting}} that works by the following priciple.
To expand $\mathcal{S}$ to a new subhypergraph, we decrease the possibility of the hyperedges in the current mode, while increase the possibility of hyperegdes with large rewards not belonging to the current mode. As a result, the possibility of hyperedges that are neighborhoods of the hyperedges in $\mathcal{S}$ with the large value of $p(e_i|\mathcal{S})$ is increased. We present an example in Fig.\ref{fig:voting} to illustrate that.

Particularly, we calculate the shifting vector $\Delta p$, such that $\mathcal{F}(\mathbf{p}^* + \Delta p)>\mathcal{F}(\mathbf{p}^*)$. According to Theorem ~\ref{thm:mode-judge}, there exist some hyperedges $e_i$, such that $\mathbf{m}(\mathbf{p}^*, \mathbf{I}_i) > \mathcal{F}(\mathbf{p}^*)$, $i\in\mathbf{G}-\mathcal{S}$. We define a direction vector $h$ as
$h_i=\mathbf{p}^*_{i}-1$ if $i\in\theta(\mathbf{p}^*)$, otherwise, $h_i = \max\{\sum_{e_j \cap e_i\neq\emptyset}p(e_j|\mathcal{S})(\mathbf{m}(\mathbf{p}^*,\mathbf{I}_i)
- \mathcal{F}(\mathbf{p}^*)), 0\}$. The above definition of $h_i$ for $i \in \theta(\mathbf{p}^*)$, decreases the possibility of $e_i$ in the current mode. However, we try to preserve the dominant seeds with a larger value of $\mathbf{p}^*_{i} - 1$, and increase the possibility of the hyperedges $e_j \in \mathbf{G} - \mathcal{S}$ that are the neighborhoods of dominant seeds of the current mode.

Assume $\mathcal{F}(h) = \eta$, then we have:
\vspace{-2mm}
\begin{align}\label{eq:time}
&\mathbf{Q}(c) = \mathcal{F}(\mathbf{p}^* + ch) - \mathcal{F}(\mathbf{p}^*) \\ \nonumber
& = \eta c^2 + 2c(p^*)^{T}\mathbf{M}h
\end{align}
We want to maximize Eq.~\eqref{eq:time}, which is the quadratic function of $c$.
Since $\Delta = 4(p^*\mathbf{M}h)^{2} > 0$, if $\eta < 0$, then
we have $c = \frac{p^*\mathbf{M}h}{\lambda}$. Otherwise, for $i \in \theta(\mathbf{p}^*)$, we have $\mathbf{p}^*_i + c(\mathbf{p}^*_i - 1) \geq 0$, then $c \leq \min_i\{\frac{p^*_i}{1 - p^*_i}\}$. Thus, $c^{\star} = \min\{\frac{p^*\mathbf{M}h}{\lambda},\min_i\{\frac{p^*_i}{1 - p^*_i}\}$, and $\Delta p = c^{\star}h$, which is the expansion vector.

We summarize the procedure of hypergraph shift in Algorithm \ref{alg:hyper-shift}.

\begin{algorithm}[hbt]
\KwIn{The hyperedge-adjacency matrix $\mathbf{M}$ of hypergraph $\mathbf{G}$, the start vector $\mathbf{p}$ (a cluster of hyperedges).}
\KwOut{The mode of hypergraph $\mathbf{G}$.}
\While{$\mathbf{p}$ is not the mode of $\mathbf{G}$ }{
Evolve $\mathbf{p}$ towards the mode of subhypergraph $\mathbf{G}_\theta(\mathbf{p})$ by Eq. (\ref{eq:replicator})\;
\eIf{ $\mathbf{p}$ is not the mode of hypergrah G}{
Expand $\mathbf{p}$ by using expansion vector $\Delta p$\;
Update $\mathbf{p}$ by mode seeking\;
}{
return\;
}
}
\caption{Hypergraph Shift Algorithm.}\label{alg:hyper-shift}
\end{algorithm}
One may wonder whether Algorithm ~\ref{alg:hyper-shift} converges, we answer this question
in theorem ~\ref{thm:conv}.
\begin{theorem}\label{thm:conv}
Algorithm ~\ref{alg:hyper-shift} is convergent.

\rm
\noindent\textbf{Proof.} The mode sequence set $\{(\mathbf{p}^*)(t)\}_{t = 1}^{\infty} \subset U$ generated by Algorithm ~\ref{alg:hyper-shift} is compact.
We construct $-\mathcal{F}(\mathbf{p})$, which is a continuous and strict decreasing function over the trajactory of sequence set. Assume the solution set is $\Gamma$, then the mode sequence generated by Algorithm ~\ref{thm:mode-judge} is closed on $U - \Gamma$. The above three conclusions are identical to the convergence conditions of Zangwill convergence theorem \cite{Zangwill}. $\hfill\blacksquare$
\end{theorem}
\section{Experimental Evaluations}\label{sec:exp}

In this section, we conduct extensive experiments to evaluate the performance of hypergraph shift.
Specific experimental setting are elaborated in each experiment.

{\bf Competitors.}
We compare our algorithm against a few closely related methods, which are introduced as follows.

For clustering evaluations, we consider the following competitors:
\begin{itemize}
\item The method proposed by Liu \etal in \cite{Cluster-ensemble}, denote by Liu \etal in follows.
\item The approach presented by Bulo \etal in \cite{Game-theo}, denote by Bulo \etal in follows.
\item Efficient hypergraph clustering \cite{EHC} (\textsf{EHC}) aims to handle the higher-order relationships among data points and seek clusters by iteratively updating the cluster membership for all nodes in parallel, and converges relatively fast.
\end{itemize}

For graph matching, we compare our method to the state-of-the-arts below:
\begin{itemize}
\item Graph shift (\textsf{GS}).
\item Two hypergraph matching methods (\textsf{TM}) \cite{Tensor-Match} and (\textsf{PM}) \cite{Pro-Match}.
\item \textsf{SC+IPFP}. The algorithm of spectral clustering \cite{Spectral} (\textsf{SC}), enhanced by the technique of integer projected fixed point \cite{IPFP}, namely \textsf{SC+IPFP} is an effective method in graph matching. Thus, it is suitable to compare our method against \textsf{SC+IPFP} in terms of graph matching.
\end{itemize}

\subsection{Clustering Analysis}

\begin{figure}[hbt]
\begin{center}
\begin{tabular}{ccc}
\hspace*{-3mm}\epsfxsize=40mm\epsfbox{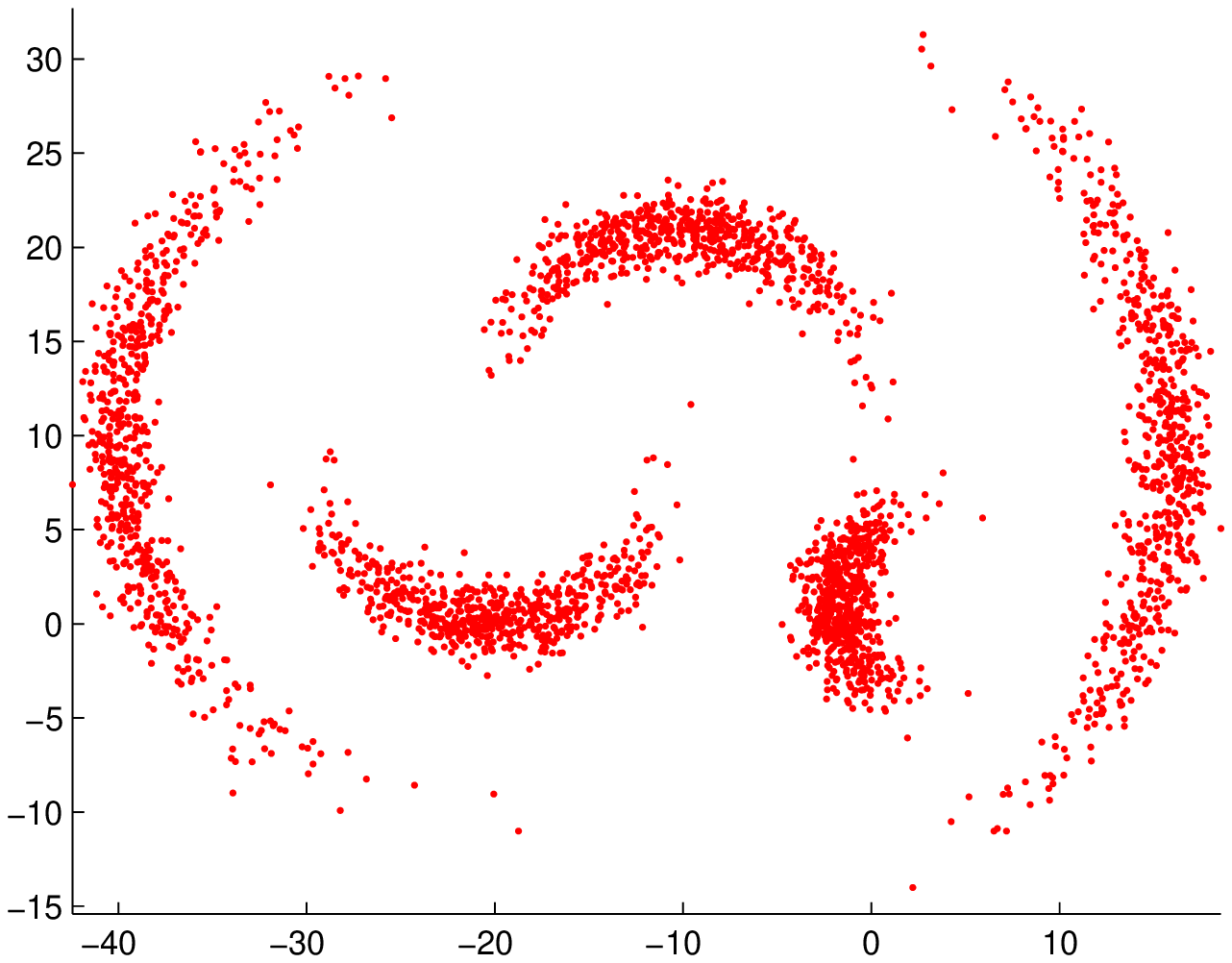}&
\hspace*{-3mm}\epsfxsize=40mm\epsfbox{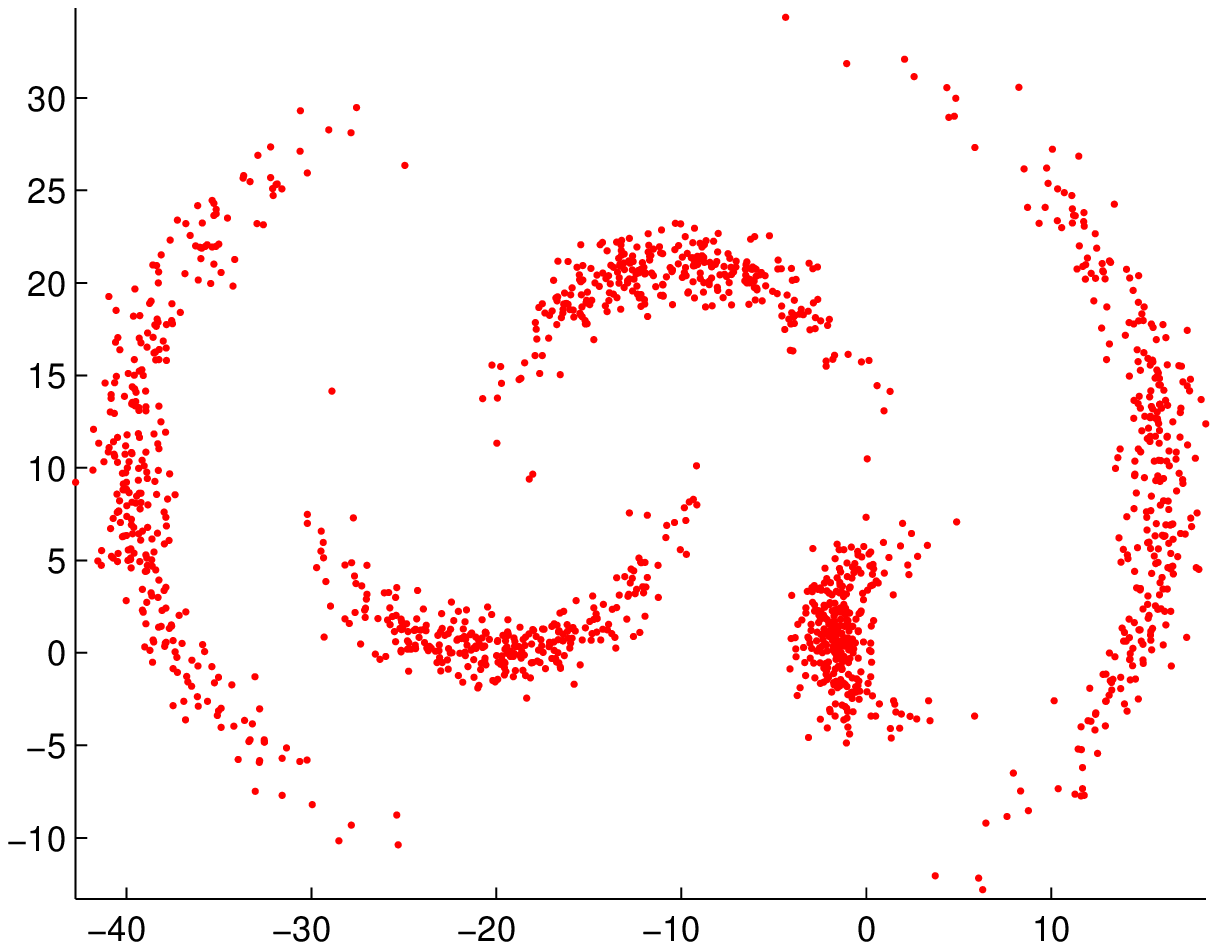}&
\hspace*{-3mm}\epsfxsize=40mm\epsfbox{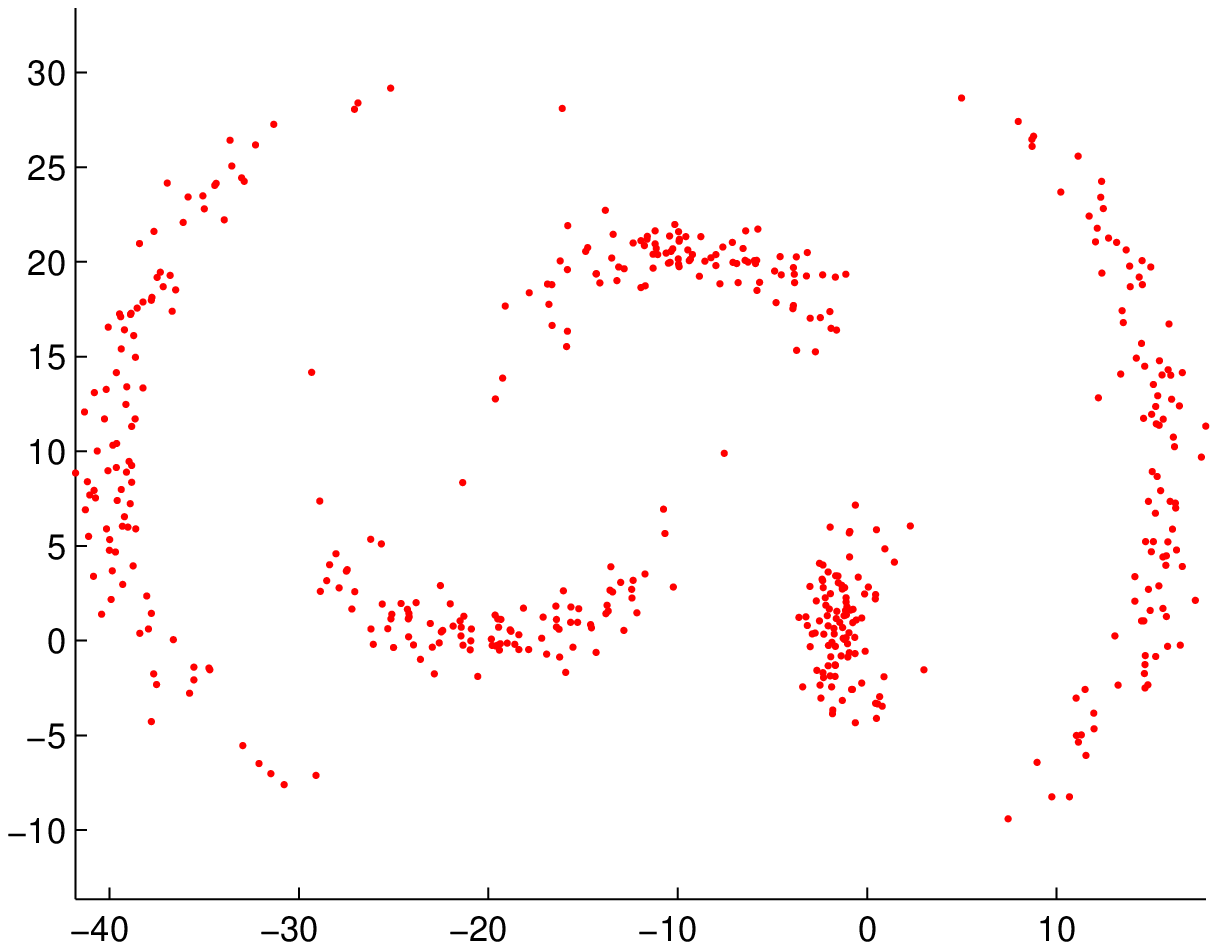}\\
(a)&(b)&(c)
\end{tabular}\\
\caption{``Five crescents'' examples with decreasing sample points from 600 pts in (a) to 300 pts in (b) and 200 pts in (c).}
\label{fig:samples}
\end{center}
\end{figure}

Consider that hypergraph shift is a natural clustering tool, and all the hyperedges shifting towards the same mode should belong to a cluster. To evaluate the clustering performance, we compare \textsf{HS} against Liu \etal, Bulo \etal and \textsf{EHC} over the data set of five crescents, as shown in Fig.\ref{fig:samples}.

\begin{figure}[hbt]%
\begin{center}
\begin{tabular}{cc}
\epsfxsize=50mm\epsfbox{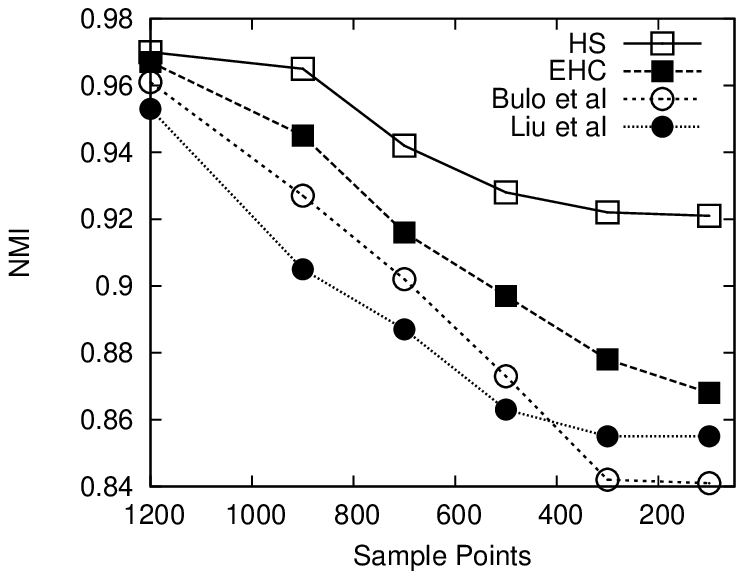}&
\epsfxsize=50mm\epsfbox{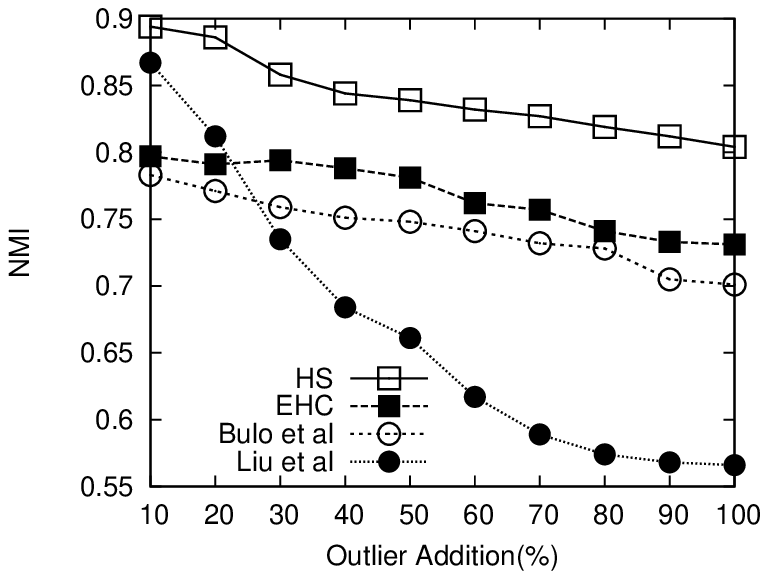}\\
(a) & (b)
\end{tabular}
\caption{Clustering comparisons on different sample sizes. We illustrate the averaged NMI with respect to the number of sample points and the ratio of outlier addition in (a) and (b), respectively.}
\label{fig:five-crescent}
\end{center}
\end{figure}

We performed extensive tests including clustering accuracy and noise robustness on five crescents gradually decreasing sampling density from 1200 pts to 100 pts. We used the standard clustering metric, normalized mutual information (NMI).
The NMI accuracy are computed for each method in Fig.\ref{fig:five-crescent} (a), with respect to decreasing sample points and increasing outliers. It shows that hypergraph shift has the best performance even in sparse data, whereas \textsf{EHC} quickly degenerates from 600 pts. The accuracies of methods in Liu \etal and Bulo \etal are inferior to \textsf{EHC}, which is consistent to the results in \cite{EHC}. To test the robustness against noises, we add Gaussian noise $\epsilon$, such that $\epsilon \sim \mathcal{N}(0,4)$, in accordance with \cite{GraphShift}, to the five crescents samples, and re-compute the NMI values. As illustrated in Fig.\ref{fig:five-crescent} (b), the three baselines of Liu \etal, Bulo \etal and \textsf{EHC} drop faster than hypergraph shift. This is because the eigenvectors required by Liu \etal are affected by all weights, no matter they are deteriorated or not; \textsf{EHC} is better than Liu \etal and Bulo \etal, however, it performs clustering by only considering the strength of affinity relationship within a hyperedge, which is not as robust against noises as the mode with high-order constraints; Hypergraph shift, in contrast, can find a dense high-order subhypergraph, which is more robust to noises.

\begin{figure}[hbt]
\begin{center}
\begin{tabular}{cc}
\epsfxsize=50mm\epsfbox{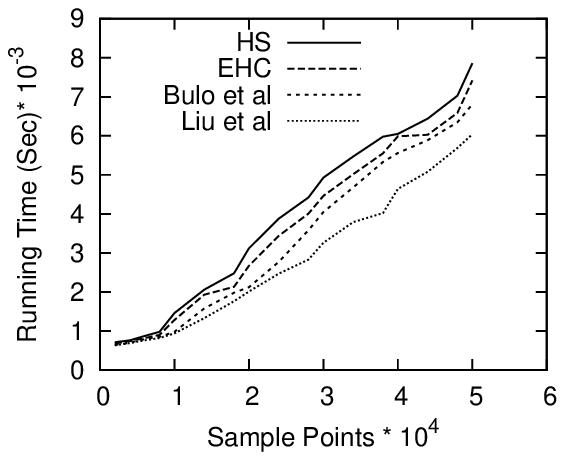}&
\epsfxsize=50mm\epsfbox{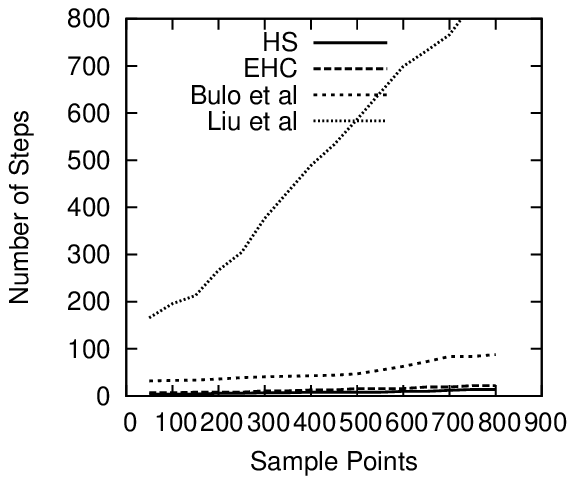}\\
(a) & (b)
\end{tabular}
\caption{The computational cost as a function of the sample size. (a) The running time against increasing number of sample points.
(b) The number of steps to convergence against increasing number of sample points.}
\label{fig:cost}
\end{center}
\end{figure}

We are interested in another important aspect: speed of convergence, under varying number of data points. In Fig.\ref{fig:cost}, we present the evaluation of the computational cost of the four methods with varying number of data points. Fig.\ref{fig:cost} (a) shows the average computational time per iteration of each method against the number of samples. We can see that the computation time per step for each method varies almost linearly with the number of data points. As expected, the least expensive method per step is Liu \etal, which performs update in sequence. And our method proceeds with expansion and dropping strategy, in the expense of more time.
However, the drawback of Liu \etal is its large iterations to convergence. In contrast, both ours and \textsf{EHC} are relatively stable w.r.t. the number of samples. Our method converges very fast, requiring on average 10 iterations. This figure experimentally show that our method, by taking larger steps towards a maximum, has significantly better speed of convergence with slightly better accuracy.

\subsection{Graph Matching}
In this part, we present some experiments on graph matching problems. We will show that this graph matching problem is identical to mode seeking on a graph with certain amount of noises and outliers. Following the experiment setup of \cite{GraphShift}, the equivalence of graph matching problem to mode seeking can be described as follows. Suppose there are two sets of feature points, $P$ and $Q$, from two images. For each point $p\in P$, we can find some similar points $q \in Q$, based on local features. Each pair of $(p,q)$ is a possible correspondence and all such pairs form the correspondence set $C=\{(p,q)| p\in P, q\in Q\}$. Then a graph $G$ is constructed based on $C$ with each vertex of $G$ representing a pair in $C$. Edge $e(v_i,v_j)$ connecting $v_i$ and $v_j$ reflects the relation between correspondence $c_i$ and $c_j$.
Due to space limitations, we refer the interested readers to \cite{GraphShift} for details.
Afterwards, the hyperedge construction and weight calculation are conducted according to our technique section.
We use the PASCAL 2012 \cite{pascal-voc-2012} database as benchmark in this evaluation.
The experiments are difficult due to the large number of outliers, that are, large amount of vertices and most of them represent incorrect correspondence, and also due to the large intra-category variations in shape present in PASCAL 2012 itself.
Under each category, we randomly select two images as a pair and calculate the matching rate by each method. We run 50 times on each category and the averaged results are report in Table \ref{tab:match-rate}. The final matching rate is averaged over rate values of all categories.

\begin{figure}[hbt]
\begin{center}
\begin{tabular}{c}
\epsfxsize=90mm\epsfbox{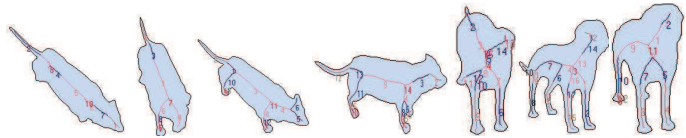}
\end{tabular}
\caption{Examples from shape matching database.}
\label{fig:dog}
\end{center}
\end{figure}

\begin{table}[hbt]
\caption{Average matching rates for the experiments on PASCAL 2012 database.}
\centering
\begin{tabular}{|c|ccccc|}
\hline\hline
&\textsf{SC+IPFP}& \textsf{GS}& \textsf{TM} & \textsf{PM} &\textsf{HS}\\
\hline
Car & 62.5\% & 60.1\% & 60.7\% & 59.2\% & \textbf{66.4}\%\\
Motorbike & 62.3\% & 60.1\% & 63.5 \% & 62.7\% & \textbf{67.3}\%\\
Person & \textbf{57.6}\% & 55.7\% & 54.2\% & 48.7\% & 53.1\%\\
Animal & 46.7\% & 49.2\% & 44.9\% & 40.3\% & \textbf{54.3}\%\\
Indoor & 30.6\% & 28.5\% & 26.6\% & 24.3\% & \textbf{36.9}\%\\
\hline\hline
All-averaged & 51.8\% & 50.7\% & 50.1\% & 47.0\% & \textbf{55.8}\%\\
\hline
\end{tabular}
\label{tab:match-rate}
\end{table}

We also conducted shape matching \cite{ShapeMatch} on the affinity data on the database from ShapeMatcher\footnote{http://www.cs.toronto.edu/~dmac/ShapeMatcher/index.html}, which contains 21 objects with 128 views for each object. A few examples of dog's shape are shown in Fig.\ref{fig:dog}.
For each shape, we compute the matching score as the affinity value using the shape matching method \cite{ShapeMatch}, thus obtain a $2688 \times 2688$ affinity matrix. We compare our method with \textsf{EHC} and \textsf{GS}. The results are shown in Table \ref{tab:affinity-data}.
Both \textsf{GS} and \textsf{HS} can specify the number of objects, however, \textsf{HS} outperforms \textsf{GS} in terms of precision due to the fact that \textsf{HS} considers high-order relationship among vertices rather than pair-wise relation.

\begin{table}[hbt]
\caption{Precision results for \textsf{EHC}, \textsf{GS} and \textsf{HS} on the shape matching affinity data.}
\centering
\begin{tabular}{cccc}
\hline\hline
&\textsf{EHC}& \textsf{GS}& \textsf{HS}\\
\hline
Objects recognized & 18 & 21 & 21\\
Precision & 72.7\% & 83.5\% & 89.82 \% \\
\hline
\end{tabular}
\label{tab:affinity-data}
\end{table}

\section{Conclusion}\label{sec:con}

In this paper, we propose a novel hypergraph shift algorithm aimed at finding the
robust graph modes by probabilistic voting strategy, which are semantically sound besides the self-cohesiveness.
Experimental studies show that our paradigm outperforms the state-of-the-art clustering and matching approaches observed from both synthetic and real-world data sets. Future work would like to incorporate the multi-view features \cite{IJCAI16},\cite{ACMMM15},\cite{SIGIR15},\cite{CIKM13}, \cite{multiview2},\cite{multiview3},\cite{multiview4},\cite{multiview5} or high-level deep representations \cite{Deep1} for improvement.\\\\

\bibliographystyle{abbrv}
\bibliography{PAKDD14}

\end{document}